\documentclass[11pt]{article}

\usepackage{colacl}
\usepackage{epsfig}
\usepackage{times}
\usepackage{lingmacros}
\usepackage{pnnamed}

\title{Never Look Back: An Alternative to Centering}

\author{{\bf Michael Strube} \\ 
IRCS -- Institute for Research in Cognitive Science \\
University of Pennsylvania \\
3401 Walnut Street, Suite 400A \\
Philadelphia PA 19104 \\
\footnotesize{\tt strube@linc.cis.upenn.edu}}

\date{}

\begin{document}

\maketitle

\begin{abstract} 
I propose a model for determining the hearer's attentional state which
depends solely on a list of salient discourse entities (S-list). The
ordering among the elements of the S-list covers also the function of
the {\em backward-looking center}\ in the centering model. The
ranking criteria for the S-list are based on the distinction between
{\em hearer-old}\ and {\em hearer-new}\ discourse entities and
incorporate preferences for inter- and intra-sentential anaphora. The
model is the basis for an algorithm which operates incrementally, word
by word.

\end{abstract}

\section{Introduction}
I propose a model for determining the hearer's attentional state in
understanding discourse. My proposal is inspired by the centering
model \citeauthor{grosz83} \shortcite{grosz83,grosz95} and draws on
the conclusions of \citeauthor{strube.acl96}'s
\shortcite{strube.acl96} approach for the ranking of the {\em
forward-looking center}\ list for German. Their approach has been
proven as the point of departure for a new model which is valid for
English as well.

The use of the centering transitions in \citeauthor{brennan87}'s
\shortcite{brennan87} algorithm prevents it from being applied
incrementally (cf.\ \textcite{kehler97}). In my approach, I propose to
replace the functions of the {\em backward-looking center}\ and the
{\em centering transitions}\ by the order among the elements of the
list of salient discourse entities (S-list). The S-list ranking
criteria define a preference for {\em hearer-old}\ over {\em
hearer-new}\ discourse entities \cite{prince81} generalizing
\citeauthor{strube.acl96}'s \shortcite{strube.acl96} approach. Because
of these ranking criteria, I can account for the difference in
salience between definite NPs (mostly hearer-old) and indefinite NPs
(mostly hearer-new).

The S-list is not a local data structure associated with individual
utterances. The S-list rather describes the attentional state of the
hearer at any given point in processing a discourse. The S-list is
generated incrementally, word by word, and used
immediately. Therefore, the S-list integrates in the simplest manner
preferences for inter- and intra-sentential anaphora, making further
specifications for processing complex sentences unnecessary.

Section \ref{sec:back} describes the centering model as the relevant
background for my proposal. In Section \ref{sec:decent}, I introduce
my model, its only data structure, the S-list, and the accompanying
algorithm. In Section \ref{sec:eval}, I compare the results of my
algorithm with the results of the centering algorithm \cite{brennan87}
with and without specifications for complex sentences
\cite{kameyama98}.

\section{A Look Back: Centering}
\label{sec:back}
The centering model describes the relation between the focus of
attention, the choices of referring expressions, and the perceived
coherence of discourse. The model has been motivated with evidence
from preferences for the antecedents of pronouns \citeauthor{grosz83}
\shortcite{grosz83,grosz95} and has been applied to pronoun resolution
(\textcite{brennan87}, inter alia, whose interpretation differs from
the original model).

The centering model itself consists of two constructs, the {\em
backward-looking center}\ and the list of {\em forward-looking
centers}, and a few rules and constraints. Each utterance $U_i$ is
assigned a list of {\em forward-looking centers}, $Cf(U_i)$, and a
unique {\em backward-looking center}, $Cb(U_i)$. A ranking imposed on
the elements of the $Cf$ reflects the assumption that the most highly
ranked element of $Cf(U_i)$ (the {\em preferred center}\ $Cp(U_i)$) is
most likely to be the $Cb(U_{i+1})$. The most highly ranked element of
$Cf(U_i)$ that is {\em realized}\ in $U_{i+1}$ (i.e., is associated
with an expression that has a valid interpretation in the underlying
semantic representation) is the $Cb(U_{i+1})$. Therefore, the ranking
on the {\em Cf}\ plays a crucial role in the model. \textcite{grosz95}
and \textcite{brennan87} use grammatical relations to rank the {\em
Cf}\ (i.e., {\em subj}\ $\prec$ {\em obj}\ $\prec$ ...) but state that other factors might
also play a role.

For their centering algorithm, \citeauthor{brennan87}
\shortcite[henceforth BFP-algorithm]{brennan87} extend the notion of
centering transition relations, which hold across adjacent utterances,
to differentiate types of shift (cf.\ Table \ref{tab:trans} taken from
\textcite{walker94}).

\begin{table}[htb]
\centering
\footnotesize
\begin{tabular}{l|c|c}
& $Cb(U_i)$ $=$ $Cb(U_{i-1})$ & $Cb(U_i)$ $\not=$ \\
& OR no $Cb(U_{i-1})$ & $Cb(U_{i-1})$ \\
\hline
$Cb(U_i)$ $=$ & & \\
$Cp(U_i)$ & \raisebox{1.3ex}[-1.3ex]{{\sc continue}} & \raisebox{1.3ex}[-1.3ex]{{\sc smooth-shift}} \\
\hline
$Cb(U_i)$ $\not=$ & & \\
$Cp(U_i)$ & \raisebox{1.3ex}[-1.3ex]{{\sc retain}} &
\raisebox{1.3ex}[-1.3ex]{{\sc rough-shift}} \\
\hline
\end{tabular}
\caption{\label{tab:trans}Transition Types}
\end{table}
\noindent
\textcite{brennan87} modify the second of two rules on center movement
and realization which were defined by \citeauthor{grosz83}
\shortcite{grosz83,grosz95}:
\begin{description}

\item[Rule 1:] If some element of $Cf(U_{i-1})$ is realized as a
pronoun in $U_i$, then so is $Cb(U_i)$.

\item[Rule 2:] Transition states are ordered. {\sc continue} is
preferred to {\sc retain} is preferred to {\sc smooth-shift} is
preferred to {\sc rough-shift}.

\end{description}
The BFP-algorithm (cf.\ \textcite{walker94}) consists of three basic
steps:

\begin{enumerate}

\item
{\sc Generate} possible {\em Cb-Cf}\ combinations.

\item
{\sc Filter} by constraints, e.g., contra-indexing, sortal predicates,
centering rules and constraints.

\item
{\sc Rank} by transition orderings.

\end{enumerate}
To illustrate this algorithm, we consider example (\ref{ex:1})
\cite{brennan87} which has two different final utterances
(\ref{ex:1}d) and (\ref{ex:1}d$^{\prime}$). Utterance (\ref{ex:1}d)
contains one pronoun, utterance (\ref{ex:1}d$^{\prime}$) two
pronouns. We look at the interpretation of (\ref{ex:1}d) and
(\ref{ex:1}d$^{\prime}$). After step 2, the algorithm has produced two
readings for each variant which are rated by the corresponding
transitions in step 3. In (\ref{ex:1}d), the pronoun {\em ``she''}\ is
resolved to {\em ``her''}\ (= Brennan) because the {\sc continue}
transition is ranked higher than {\sc smooth-shift} in the second
reading. In (\ref{ex:1}d$^{\prime}$), the pronoun {\em ``she''}\ is
resolved to {\em ``Friedman''}\ because {\sc smooth-shift} is preferred
over {\sc rough-shift}.

\eenumsentence{\label{ex:1}
\small

\item[a.]
Brennan drives an Alfa Romeo.

\item[b.]
{\bf She} drives too fast.

\item[c.]
Friedman races {\bf her} on weekends.

\item[d.]
{\bf She} goes to Laguna Seca.

\item[d.$^\prime$]
{\bf She} often beats {\bf her}.

}

\section{An Alternative to Centering}
\label{sec:decent}
\subsection{The Model}
\label{subsec:model}
The realization and the structure of my model departs significantly
from the centering model:
\begin{itemize}

\item
\vspace{-0.5ex}
The model consists of one construct with one operation: the list of
salient discourse entities (S-list) with an insertion operation. 

\item
\vspace{-0.5ex}
The S-list describes the attentional state of the hearer at any given
point in processing a discourse. 

\item
\vspace{-0.5ex}
The S-list contains some (not necessarily all) discourse entities
which are realized in the current and the previous utterance.

\item
\vspace{-0.5ex}
The elements of the S-list are ranked according to their information
status. The order among the elements provides directly the preference
for the interpretation of anaphoric expressions.
\vspace{-2.5ex}
\end{itemize}
In contrast to the centering model, my model does not need a construct which
looks back; it does not need transitions and transition ranking
criteria. %
%
Instead of using the {\em Cb}\ to account for local coherence, in my
model this is achieved by comparing the first element of the S-list
with the preceding state.

\subsection{S-List Ranking}
\label{subsec:rank}
\begin{figure}[htb]
\centering
\mbox{
\epsfxsize=\columnwidth
\epsfbox{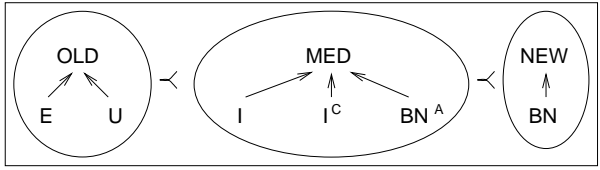}
}
\caption{S-list Ranking and Familiarity}
\label{fig:fam}
\end{figure}

\textcite{strube.acl96} rank the {\em Cf}\ according to the
information status of discourse entities. I here generalize these
ranking criteria by redefining them in \citeauthor{prince81}'s
\shortcite{prince81,prince92} terms. I distinguish between three
different sets of expressions, {\em hearer-old discourse entities}\
(OLD), {\em mediated discourse entities}\ (MED), and {\em hearer-new
discourse entities}\ (NEW). These sets consist of the elements of
Prince's {\em familiarity scale}\ \cite[p.245]{prince81}. OLD consists
of {\em evoked}\ (E) and {\em unused}\ (U) discourse entities while
NEW consists of {\em brand-new} (BN) discourse entities. MED consists
of {\em inferrables}\ (I), {\em containing inferrables}\ (I$^C$) and
{\em anchored brand-new}\ (BN$^A$) discourse entities. These discourse
entities are {\em discourse-new}\ but {\em mediated} by some {\em
hearer-old}\ discourse entity (cf.\ Figure \ref{fig:fam}). I do not
assume any difference between the elements of each set with respect to
their information status. E.g., evoked and unused discourse entities
have the same information status because both belong to OLD.

For an operationalization of Prince's terms, I stipulate that {\em
evoked}\ discourse entitites are co-referring expressions (pronominal
and nominal anaphora, previously mentioned proper names, relative
pronouns, appositives). {\em Unused}\ discourse entities are proper
names and titles. In texts, {\em brand-new}\ proper names are usually
accompanied by a relative clause or an appositive which relates them
to the hearer's knowledge. The corresponding discourse entity is {\em
evoked}\ only after this elaboration. Whenever these linguistic
devices are missing, proper names are treated as {\em unused}%
\footnote{For examples of brand-new proper names and their
introduction cf., e.g., the ``obituaries'' section of the {\em New
York Times}.
}. I restrict {\em inferrables}\ to the particular subset defined by
\textcite{hahn.ecai96}. {\em Anchored brand-new}\ discourse entities
require that the anchor is either {\em evoked}\ or {\em unused}.

I assume the following conventions for the ranking constraints on the
elements of the S-list. The 3-tuple $(x, utt_x, pos_x)$ denotes a
discourse entity $x$ which is {\em evoked}\ in utterance $utt_x$ at
the text position $pos_x$. With respect to any two discourse entities
$(x, utt_x, pos_x)$ and $(y, utt_y, pos_y)$, $utt_x$ and $utt_y$
specifying the current utterance $U_i$ or the preceding utterance
$U_{i-1}$, I set up the following ordering constraints on elements in
the S-list (Table \ref{tab:rank})%
\footnote{The relations $\succ$ and = indicate that the utterance
containing {\em x}\ follows ($\succ$) the utterance containing {\em
y}\ or that {\em x}\ and {\em y}\ are elements of the same utterance
(=).
}. For any state of the processor/hearer, the ordering of discourse
entities in the S-list that can be derived from the ordering
constraints (1) to (3) is denoted by the precedence relation $\prec$.

\begin{table}[htb]
\centering
\footnotesize
\begin{tabular}{|ll|}

\hline

(1) & If $x \in$ OLD and $y \in$ MED, then $x \prec y$. \\
& If $x \in$ OLD and $y \in$ NEW, then $x \prec y$. \\
& If $x \in$ MED and $y \in$ NEW, then $x \prec y$. \\
& \\[-1ex]

(2) & If $x, y \in$ OLD, or $x, y \in$ MED, or $x, y \in$ NEW, \\
& then if $utt_x \succ utt_y$, then $x \prec y$, \\
& \qquad if $utt_x = utt_y$ and $pos_x < pos_y$, then $x \prec y$.  \\
%
%
\hline

\end{tabular}
\caption{Ranking Constraints on the S-list}
\label{tab:rank}
\end{table}

Summarizing Table \ref{tab:rank}, I state the following preference
ranking for discourse entities in $U_i$ and $U_{i-1}$: {\em
hearer-old}\ discourse entities in $U_i$, {\em hearer-old}\ discourse
entities in $U_{i-1}$, {\em mediated}\ discourse entities in $U_i$,
{\em mediated}\ discourse entities in $U_{i-1}$, {\em hearer-new}\
discourse entities in $U_i$, {\em hearer-new}\ discourse entities in
$U_{i-1}$. By making the distinction in (2) between discourse entities
in $U_i$ and discourse entities in $U_{i-1}$, I am able to deal with
intra-sentential anaphora. There is no need for further specifications
for complex sentences. A finer grained ordering is achieved by
ranking discourse entities within each of the sets according to
their text position.


\subsection{The Algorithm}
\label{subsec:algo}
Anaphora resolution is performed with a simple look-up in the S-list%
\footnote{The S-list consists of referring expressions which are
specified for text position, agreement, sortal information, and
information status. Coordinated NPs are collected in a set. The S-list
does not contain predicative NPs, pleonastic {\em ``it''}, and any
elements of direct speech enclosed in double quotes.
}. The elements of the S-list are tested in the given order until one
test succeeds. Just after an anaphoric expression is resolved, the
S-list is updated. The algorithm processes a text from left to
right (the unit of processing is the word):

\begin{enumerate}

\item
\vspace{-0,5ex}
If a referring expression is encountered,

\begin{enumerate}

\item
\vspace{-0,5ex}
if it is a pronoun, test the elements of the S-list in the given order
until the test succeeds%
\footnote{The test for pronominal anaphora involves checking agreement
criteria, binding and sortal constraints.%
%
};

\item
\vspace{-0,5ex}
update S-list; the position of the referring expression under
consideration is determined by the S-list-ranking criteria which are
used as an insertion algorithm.

\end{enumerate}

\item
\vspace{-0,5ex}
If the analysis of utterance {\em U}%
\footnote{I here define that an utterance is a sentence.
} is finished, remove all
discourse entities from the S-list, which are not {\em realized}\ in
{\em U}.
\vspace{-0,5ex}

\end{enumerate}

The analysis for example (\ref{ex:1}) is given in Table
\ref{tab:decent-ex1}%
\footnote{In the following Tables, discourse entities are represented
by {\sc SmallCaps}, while the corresponding surface expression appears
on the right side of the colon. Discourse entitites are annotated with
their information status. An ``$\epsilon$'' indicates an elliptical
NP.
}. I show only these steps which are of interest for the computation
of the S-list and the pronoun resolution. The preferences for pronouns
(in bold font) are given by the S-list immediately above them. The
pronoun {\em ``she''}\ in (\ref{ex:1}b) is resolved to the first
element of the S-list. When the pronoun {\em ``her''}\ in
(\ref{ex:1}c) is encountered, {\sc Friedman} is the first element of
the S-list since {\sc Friedman} is {\em unused} and in the current
utterance. Because of binding restrictions, {\em ``her''}\ cannot be
resolved to {\sc Friedman} but to the second element, {\sc Brennan}. In
both (\ref{ex:1}d) and (\ref{ex:1}d$^\prime$) the pronoun {\em
``she''}\ is resolved to {\sc Friedman}.

\begin{table}[htb]

\centering

\begin{footnotesize}

\begin{tabular}{|c|ll|}

\hline
(\ref{ex:1}a) & \multicolumn{2}{|l|}{Brennan drives an Alfa Romeo} \\
& & S: [{\sc Brennan}$_U$: Brennan, \\
& & \hspace{1.6em}{\sc Alfa Romeo}$_{BN}$: Alfa Romeo] \\

\hline
(\ref{ex:1}b) & \multicolumn{2}{|l|}{{\bf She} drives too fast.} \\
& & S: [{\sc Brennan}$_E$: she] \\

\hline
(\ref{ex:1}c) & \multicolumn{2}{|l|}{Friedman} \\
& & S: [{\sc Friedman}$_U$: Friedman, {\sc Brennan}$_E$: she] \\

& \multicolumn{2}{|l|}{races {\bf her} on weekends.} \\
& & S: [{\sc Friedman}$_U$: Friedman, {\sc Brennan}$_E$: her] \\

\hline
(\ref{ex:1}d) & \multicolumn{2}{|l|}{{\bf She} drives to Laguna Seca.} \\
& & S: [{\sc Friedman}$_E$: she, \\
& & \hspace{1.6em}{\sc Laguna Seca}$_U$: Laguna Seca] \\

\hline
(\ref{ex:1}d$^{\prime}$) & \multicolumn{2}{|l|}{{\bf She}} \\
& & S: [{\sc Friedman}$_E$: she, {\sc Brennan}$_E$: her] \\

& \multicolumn{2}{|l|}{often beats {\bf her}.} \\
& & S: [{\sc Friedman}$_E$: she, {\sc Brennan}$_E$: her] \\

\hline

\end{tabular}

\end{footnotesize}
\caption{Analysis for (\ref{ex:1})}
\label{tab:decent-ex1}
\end{table}

The difference between my algorithm and the BFP-algorithm becomes
clearer when the {\em unused}\ discourse entity {\em ``Friedman''}\ is
replaced by a {\em brand-new}\ discourse entity, e.g., {\em ``a
professional driver''}%
\footnote{I owe this variant Andrew Kehler. -- This example can
misdirect readers because the phrase {\em ``a professional driver''}\
is assigned the ``default'' gender masculine. Anyway, this example --
like the original example -- seems not to be felicitous English and
has only illustrative character.
} (cf.\ example (\ref{ex:2})). In the BFP-algorithm, the ranking of
the {\em Cf}-list depends on grammatical roles. Hence, {\sc Driver} is
ranked higher than {\sc Brennan} in the {\em Cf}(\ref{ex:2}c). In
(\ref{ex:2}d), the pronoun {\em ``she''}\ is resolved to {\sc Brennan}
because of the preference for {\sc continue} over {\sc retain}. In
(\ref{ex:2}d$^{\prime}$), {\em ``she''}\ is resolved to {\sc Driver}
because {\sc smooth-shift} is preferred over {\sc rough-shift}. In my
algorithm, at the end of (\ref{ex:2}c) the {\em evoked}\ phrase {\em
``her''}\ is ranked higher than the {\em brand-new}\ phrase {\em ``a
professional driver''} (cf.\ Table \ref{tab:decent-ex2}). In both
(\ref{ex:2}d) and (\ref{ex:2}d$^{\prime}$) the pronoun {\em ``she''}\
is resolved to {\sc Brennan}.

\eenumsentence{\label{ex:2}
\small

\item[a.]\vspace{-0,5ex}
Brennan drives an Alfa Romeo.

\item[b.]\vspace{-0,5ex}
{\bf She} drives too fast.

\item[c.]\vspace{-0,5ex}
A professional driver races {\bf her} on weekends.

\item[d.]\vspace{-0,5ex}
{\bf She} goes to Laguna Seca.

\item[d.$^\prime$]\vspace{-0,5ex}
{\bf She} often beats {\bf her}.
\vspace{-0,5ex}
}

\begin{table}[htb]

\centering

\begin{footnotesize}

\begin{tabular}{|c|ll|}

\hline
(\ref{ex:2}a) & \multicolumn{2}{|l|}{Brennan drives an Alfa Romeo} \\
& & S: [{\sc Brennan}$_U$: Brennan, \\
& & \hspace{1.6em}{\sc Alfa Romeo}$_{BN}$: Alfa Romeo] \\

\hline
(\ref{ex:2}b) & \multicolumn{2}{|l|}{{\bf She} drives too fast.} \\
& & S: [{\sc Brennan}$_E$: she] \\

\hline
(\ref{ex:2}c) & \multicolumn{2}{|l|}{A professional driver} \\
& & S: [{\sc Brennan}$_E$: she, {\sc Driver}$_{BN}$: Driver] \\

& \multicolumn{2}{|l|}{races {\bf her} on weekends.} \\
& & S: [{\sc Brennan}$_E$: her, {\sc Driver}$_{BN}$: Driver] \\

\hline
(\ref{ex:2}d) & \multicolumn{2}{|l|}{{\bf She} drives to Laguna Seca.} \\
& & S: [{\sc Brennan}$_E$: she, \\
& & \hspace{1.6em}{\sc Laguna Seca}$_U$: Laguna Seca] \\

\hline
(\ref{ex:2}d$^{\prime}$) & \multicolumn{2}{|l|}{{\bf She}} \\
& & S: [{\sc Brennan}$_E$: she, {\sc Driver}$_{BN}$: Driver] \\

& \multicolumn{2}{|l|}{often beats {\bf her}.} \\
& & S: [{\sc Brennan}$_E$: she, {\sc Driver$_E$:} her] \\

\hline

\end{tabular}

\end{footnotesize}
\caption{Analysis for (\ref{ex:2})}
\label{tab:decent-ex2}
\end{table}
\begin{table*}[htb]

\centering

\begin{footnotesize}

\begin{tabular}{|c|ll|}

\hline
(\ref{ex:3}a) & \multicolumn{2}{|l|}{A judge} \\
& & S: [{\sc Judge}$_{BN}$: judge] \\

& \multicolumn{2}{|l|}{ordered that Mr.\ Curtis} \\
& & S: [{\sc Curtis}$_E$: Mr.\ Curtis, {\sc Judge}$_{BN}$: judge] \\

& \multicolumn{2}{|l|}{be released, but $\epsilon$} \\
& & S: [{\sc Curtis}$_E$: Mr.\ Curtis, {\sc Judge}$_E$: $\epsilon$] \\

& \multicolumn{2}{|l|}{agreed with a request} \\
& & S: [{\sc Curtis}$_E$: Mr.\ Curtis, {\sc Judge}$_E$: $\epsilon$,
{\sc Request}$_{BN}$: request] \\

& \multicolumn{2}{|l|}{from prosecutors} \\
& & S: [{\sc Curtis}$_E$: Mr.\ Curtis, {\sc Judge}$_E$: $\epsilon$,
{\sc Request}$_{BN}$: request, {\sc Prosecutors}$_{BN}$: prosecutors] \\

& \multicolumn{2}{|l|}{that {\bf he}} \\
& & S: [{\sc Curtis}$_E$: he, {\sc Judge}$_E$: $\epsilon$,
{\sc Request}$_{BN}$: request, {\sc Prosecutors}$_{BN}$: prosecutors] \\

& \multicolumn{2}{|l|}{be re-examined each year} \\
& & S: [{\sc Curtis}$_E$: he, {\sc Judge}$_E$: $\epsilon$,
{\sc Request}$_{BN}$: request, {\sc Prosecutors}$_{BN}$: prosecutors, {\sc
Year}$_{BN}$: year] \\

& \multicolumn{2}{|l|}{to see if {\bf his}} \\
& & S: [{\sc Curtis}$_E$: his, {\sc Judge}$_E$: $\epsilon$,
{\sc Request}$_{BN}$: request, {\sc Prosecutors}$_{BN}$: prosecutors, {\sc
Year}$_{BN}$: year] \\

& \multicolumn{2}{|l|}{condition} \\
& & S: [{\sc Curtis}$_E$: his, {\sc Judge}$_E$: $\epsilon$, {\sc
Condition}$_{{BN}^A}$: condition, {\sc Request}$_{BN}$: request, {\sc
Prosecutors}$_{BN}$: prosec.] \\

& \multicolumn{2}{|l|}{has improved.} \\
& & S: [{\sc Curtis}$_E$: his, {\sc Judge}$_E$: $\epsilon$, {\sc
Condition}$_{{BN}^A}$: condition, {\sc Request}$_{BN}$: request, {\sc
Prosecutors}$_{BN}$: prosec.] \\

\hline
(\ref{ex:3}b) & \multicolumn{2}{|l|}{But authorities lost contact with
Mr.\ Curtis after the Connecticut Supreme Court ruled in 1990 that the
judge had} \\

& \multicolumn{2}{|l|}{erred, and that prosecutors had no right} \\

& & S: [{\sc Curtis}$_E$: his, {\sc CS Court}$_U$: CS Court, {\sc
Judge}$_E$: judge, {\sc Condition}$_{BN^A}$: condition, {\sc
Auth.}$_{BN}$: auth.] \\

& \multicolumn{2}{|l|}{to re-examine {\bf him}.} \\
& & S: [{\sc Curtis}$_E$: him, {\sc CS Court}$_U$: CS Court, {\sc
Judge}$_E$: judge, {\sc Condition}$_{BN^A}$: condition, {\sc
Auth.}$_{BN}$: auth.] \\

\hline
(\ref{ex:3}c) & \multicolumn{2}{|l|}{John Smirga, the assistant
state's attorney in charge of the original case, said last week} \\
& & S: [{\sc Smirga}$_E$: attorney, {\sc Case}$_E$: case, {\sc
Curtis}$_E$: him, {\sc CS Court}$_U$: CS Court, {\sc Judge}$_E$: judge
] \\

& \multicolumn{2}{|l|}{that {\bf he} had doubts about the psychiatric
reports that said Mr.\ Curtis would never improve.} \\
& & S: [{\sc Smirga}$_E$: he, {\sc Case}$_E$: case, {\sc Reports}$_E$:
reports, {\sc Curtis}$_E$: Mr.\ Curtis, {\sc Doubts}$_{BN}$: doubts] \\

\hline

\end{tabular}

\end{footnotesize}
\vspace{-0,5ex}
\caption{Analysis for (\ref{ex:3})}
\label{tab:decent-ex3}
\vspace{-1ex}
\end{table*}

Example (\ref{ex:3})%
\footnote{In: {\em The New York Times}. Dec.\ 7, 1997, p.A48 (``Shot
in head, suspect goes free, then to college'').
} illustrates how the preferences for intra- and inter-sentential
anaphora interact with the information status of discourse entitites
(Table \ref{tab:decent-ex3}). Sentence (\ref{ex:3}a) starts a new
discourse segment. The phrase {\em ``a judge''}\ is {\em brand-new}.
{\em ``Mr.\ Curtis''}\ is mentioned several times before in the text,
Hence, the discourse entity {\sc Curtis}\ is {\em
evoked}\ and ranked higher than the discourse entity {\sc
Judge}. In the next step, the ellipsis refers to {\sc Judge} which is
evoked then. The nouns {\em ``request''}\ and {\em ``prosecutors''}\
are {\em brand-new}%
\footnote{I restrict {\em inferrables}\ to the cases specified by
\textcite{hahn.ecai96}. Therefore {\em ``prosecutors''}\ is {\em
brand-new} (cf.\ \textcite{prince92} for a discussion of the form of
inferrables).
}. The pronoun {\em ``he''}\ and the possessive pronoun {\em ``his''}\
are resolved to {\sc Curtis}. {\em ``Condition''}\ is {\em brand-new}\
but anchored by the possessive pronoun. For (\ref{ex:3}b) and
(\ref{ex:3}c) I show only the steps immediately before the pronouns
are resolved. In (\ref{ex:3}b) both {\em ``Mr.\ Curtis''}\ and {\em
``the judge''}\ are {\em evoked}. However, {\em ``Mr.\ Curtis''}\ is
the left-most {\em evoked}\ phrase in this sentence and therefore the
most preferred antecedent for the pronoun {\em ``him''}. For my
experiments I restricted the length of the S-list to five
elements. Therefore {\em ``prosecutors''}\ in (\ref{ex:3}b) is not
contained in the S-list. The discourse entity {\sc Smirga} is
introduced in (\ref{ex:3}c). It becomes {\em evoked}\ after the
appositive. Hence {\sc Smirga} is the most preferred antecedent for
the pronoun {\em ``he''}.

\eenumsentence{\label{ex:3}
\small

\item[a.]\vspace{-0,5ex} 
A judge ordered that Mr.\ Curtis be released, but $\epsilon$ agreed
with a request from prosecutors that {\bf he} be re-examined each year
to see if {\bf his} condition has improved.

\item[b.]\vspace{-0,5ex}
But authorities lost contact with Mr.\ Curtis after the Connecticut
Supreme Court ruled in 1990 that the judge had erred, and that
prosecutors had no right to re-examine {\bf him}.

\item[c.]\vspace{-0,5ex}
John Smirga, the assistant state's attorney in charge of the original
case, said last week that {\bf he} always had doubts about the psychiatric
reports that said Mr.\ Curtis would never improve.
\vspace{-0,5ex}
}

\section{Some Empirical Data}
\label{sec:eval}
In the first experiment, I compare my algorithm with the
BFP-algorithm which was in a second experiment extended by the
constraints for complex sentences as described by
\textcite{kameyama98}.

\paragraph{Method.}
I use the following guidelines for the hand-simulated analysis
\cite{walker89}. I do not assume any world knowledge as part of the
anaphora resolution process. Only agreement criteria, binding and
sortal constraints are applied. I do not account for false positives
and error chains. Following \textcite{walker89}, a segment is defined
as a paragraph unless its first sentence has a pronoun in subject
position or a pronoun where none of the preceding sentence-internal
noun phrases matches its syntactic features. At the beginning of
a segment, anaphora resolution is preferentially performed within the
same utterance. My algorithm starts with an empty S-list at the
beginning of a segment.

The basic unit for which the centering data structures are generated
is the utterance {\em U}. For the BFP-algorithm, I define {\em U}\
as a simple sentence, a complex sentence, or each full clause of a
compound sentence. \citeauthor{kameyama98}'s \shortcite{kameyama98}
intra-sentential centering operates at the clause level. While tensed
clauses are defined as utterances on their own, untensed clauses are
processed with the main clause, so that the {\em Cf}-list of the main
clause contains the elements of the untensed embedded
clause. \citeauthor{kameyama98} distinguishes for tensed clauses
further between sequential and hierarchical centering. Except for
reported speech (embedded and inaccessible to the superordinate
level), non-report complements, and relative clauses (both embedded
but accessible to the superordinate level; less salient than the
higher levels), all other types of tensed clauses build a chain of
utterances on the same level.

According to the preference for inter-sentential candidates in the
centering model, I define the following anaphora resolution strategy
for the BFP-algorithm: (1) Test elements of $U_{i-1}$. (2) Test
elements of $U_i$ left-to-right. (3) Test elements of $Cf(U_{i-2})$,
$Cf(U_{i-3})$, ...  In my algorithm steps (1) and (2) fall
together. (3) is performed using previous states of the system.

\paragraph{Results.}
The test set consisted of the beginnings of three short stories by
Hemingway (2785 words, 153 sentences) and three articles from the {\em
New York Times}\ (4546 words, 233 sentences). The results of my
experiments are given in Table \ref{tab:eval}. The first row gives the
number of personal and possessive pronouns. The remainder of the Table
shows the results for the BFP-algorithm, for the BFP-algorithm
extended by \citeauthor{kameyama98}'s intra-sentential specifications,
and for my algorithm. The overall error rate of each approach is given
in the rows marked with {\em wrong}. The rows marked with {\em wrong
(strat.)}\ give the numbers of errors directly produced by the
algorithms' strategy, the rows marked with {\em wrong (ambig.)}\ the
number of analyses with ambiguities generated by the BFP-algorithm (my
approach does not generate ambiguities). The rows marked with {\em
wrong (intra)}\ give the number of errors caused by (missing)
specifications for intra-sentential anaphora. Since my algorithm
integrates the specifications for intra-sentential anaphora, I count
these errors as strategic errors. The rows marked with {\em wrong
(chain)}\ give the numbers of errors contained in error chains. The
rows marked with {\em wrong (other)} give the numbers of the remaining
errors (consisting of pronouns with split antecedents, errors because
of segment boundaries, and missing specifications for event anaphora).

\begin{table}[htb]
\centering
\footnotesize

\begin{tabular}{|l|l|cc|c|}

\hline
\multicolumn{2}{|l|}{} & {\bf Hem.} & {\bf NYT} & {\bf $\Sigma$} \\

\hline
\multicolumn{2}{|l|}{{\bf Pron.\ and Poss.\ Pron.}} & {\bf 274} & {\bf
302} & {\bf 576} \\
\hline
& {\bf Correct} & {\bf 189} & {\bf 231} & {\bf 420} \\
& {\bf Wrong} & {\bf 85} & {\bf 71} & {\bf 156} \\
\cline{2-5}
& Wrong (strat.) & 14 & 2 & 16 \\
{\bf BFP-Algo.} & Wrong (ambig.) & 9 & 15 & 24 \\
& Wrong (intra) & 17 & 13 & 30 \\
& Wrong (chain) & 29 & 32 & 61 \\
& Wrong (other) & 16 & 9 & 25 \\

\hline
& {\bf Correct} & {\bf 193} & {\bf 245} & {\bf 438} \\
& {\bf Wrong} & {\bf 81} & {\bf 57} & {\bf 138} \\
\cline{2-5}
& Wrong (strat.) & 3 & 0 & 3 \\
{\bf BFP/Kam.} & Wrong (ambig.) & 17 & 8 & 25 \\
& Wrong (intra) & 17 & 27 & 44 \\
& Wrong (chain) & 29 & 15 & 44 \\
& Wrong (other) & 15 & 7 & 22 \\

\hline
& {\bf Correct} & {\bf 217} & {\bf 275} & {\bf 492} \\
& {\bf Wrong} & {\bf 57} & {\bf 27} & {\bf 84} \\
\cline{2-5}
{\bf My Algo.} & Wrong (strat.) & 21 & 12 & 33 \\
& Wrong (chain) & 22 & 9 & 31 \\
& Wrong (other) & 14 & 6 & 20 \\
\hline

\end{tabular}
\caption{Evaluation Results}
\label{tab:eval}
\end{table}

\paragraph{Interpretation.}
The results of my experiments showed not only that my algorithm
performed better than the centering approaches but also revealed
insight in the interaction between inter- and intra-sentential
preferences for anaphoric antecedents. \citeauthor{kameyama98}'s
specifications reduce the complexity in that the {\em Cf}-lists in
general are shorter after splitting up a sentence into
clauses. Therefore, the BFP-algorithm combined with her specifications
has almost no strategic errors while the number of ambiguities remains
constant. But this benefit is achieved at the expense of more errors
caused by the intra-sentential specifications. These errors occur in
cases like example (\ref{ex:3}), in which \citeauthor{kameyama98}'s
intra-sentential strategy makes the correct antecedent less salient,
indicating that a clause-based approach is too fine-grained and that
the hierarchical syntactical structure as assumed by
\citeauthor{kameyama98} does not have a great impact on anaphora
resolution.

I noted, too, that the BFP-algorithm can generate ambiguous readings
for $U_i$ %
when the pronoun in $U_i$ does not co-specify the $Cb(U_{i-1})$. In
cases, where the $C_f(U_{i-1})$ contains more than one possible
antecedent for the pronoun, several ambiguous readings with the same
transitions are generated. An example%
\footnote{In: Ernest Hemingway. {\em Up in Michigan.}\ In. {\em The
Complete Short Stories of Ernest Hemingway.}\ New York: Charles
Scribner's Sons, 1987, p.60.
}: There is no $Cb(\ref{ex:5}a)$ because no
element of the preceding utterance is realized in (\ref{ex:5}a). The
pronoun {\em ``them''}\ in (\ref{ex:5}b) co-specifies {\em ``deer''}\
but the BFP-algorithm generates two readings both of which are marked
by a {\sc Retain} transition.
\eenumsentence{\label{ex:5}
\small

\item[a.]
\vspace{-0,4ex}
Jim pulled the burlap sacks off the deer

\item[b.]
\vspace{-0,7ex}
and Liz looked at {\bf them}. 
\vspace{-0,4ex}
}
In general, the strength of the centering model is that it is possible
to use the $Cb(U_{i-1})$ as the most preferred antecedent for a
pronoun in $U_i$. In my model this effect is achieved by the
preference for hearer-old discourse entities. Whenever this preference
is misleading both approaches give wrong results. Since the {\em Cb}\
is defined strictly local while hearer-old discourse entities are
defined global, my model produces less errors. In my model the
preference is available immediately while the BFP-algorithm can use
its preference not before the second utterance has been processed. The
more global definition of hearer-old discourse entities leads also to
shorter error chains. -- However, the test set is too small to draw
final conclusions, but at least for the texts analyzed the preference
for {\em hearer-old}\ discourse entities is more appropriate than the
preference given by the BFP- algorithm.

\section{Comparison to Related Approaches}
\citeauthor{kameyama98}'s \shortcite{kameyama98} version of centering
also omits the {\em centering transitions}. But she uses the {\em Cb}\ and
a ranking over simplified transitions preventing the incremental
application of her model.

The focus model \cite{sidner83,suri94} accounts for {\em evoked}\
discourse entities explicitly because it uses the discourse focus,
which is determined by a successful anaphora resolution. Incremental
processing is not a topic of these papers.

Even models which use salience measures for determining the
antecedents of pronoun use the concept of {\em evoked}\ discourse
entities. \textcite{hajicova92} assign the highest value to an evoked
discourse entity. Also \textcite{lappin94}, who give the subject of
the current sentence the highest weight, have an implicit notion of
{\em evokedness}. The salience weight degrades from one sentence to
another by a factor of two which implies that a repeatedly mentioned
discourse entity gets a higher weight than a {\em brand-new}\ subject.

\section{Conclusions}
\label{sec:conc}
In this paper, I proposed a model for determining the hearer's
attentional state which is based on the distinction between {\em
hearer-old}\ and {\em hearer-new}\ discourse entities. I showed that
my model, though it omits the {\em backward-looking center}\ and
the {\em centering transitions}, does not lose any of the predictive
power of the centering model with respect to anaphora resolution. In
contrast to the centering model, my model includes a treatment for
intra-sentential anaphora and is sufficiently well specified to be
applied to real texts. Its incremental character seems to be an answer
to the question \textcite{kehler97} recently raised. Furthermore, it
neither has the problem of inconsistency \citeauthor{kehler97}
mentioned with respect to the BFP-algorithm nor does it generate
unnecessary ambiguities.

Future work will address whether the text position, which is the
weakest grammatical concept, is sufficient for the order of the
elements of the S-list at the second layer of my ranking
constraints. I will also try to extend my model for the analysis
of definite noun phrases for which it is necessary to integrate
it into a more global model of discourse processing.

\paragraph{Acknowledgments:}
This work has been funded by a post-doctoral grant from DFG (Str
545/1-1) \linebreak and is supported by a post-doctoral fellowship
award from IRCS. I would like to thank Nobo Komagata, Rashmi Prasad,
and Matthew Stone who commented on earlier drafts of this paper. I am
grateful for valuable comments by Barbara Grosz, Udo Hahn, Aravind
Joshi, Lauri Karttunen, Andrew Kehler, Ellen Prince, and Bonnie
Webber.

\smallbibliography{\footnotesize}

\bibliographystyle{pnnamedabbrv}

\begin{thebibliography}{}

\bibitem[\protect\citeauthoryear{Brennan et~al.}{1987}]{brennan87}
Brennan,~S.~E., M.~W. Friedman {\&} C.~J. Pollard (1987).
\newblock A centering approach to pronouns.
\newblock In {\em Proc. of the 25{\/$^{th}$} Annual Meeting of the Association
  for Computational Linguistics; Stanford, Cal., 6--9 July 1987}, pp. 155--162.

\bibitem[\protect\citeauthoryear{Grosz et~al.}{1983}]{grosz83}
Grosz,~B.~J., A.~K. Joshi {\&} S.~Weinstein (1983).
\newblock Providing a unified account of definite noun phrases in discourse.
\newblock In {\em Proc. of the 21{\/$^{st}$} Annual Meeting of the Association
  for Computational Linguistics; Cambridge, Mass., 15--17 June 1983}, pp.
  44--50.

\bibitem[\protect\citeauthoryear{Grosz et~al.}{1995}]{grosz95}
Grosz,~B.~J., A.~K. Joshi {\&} S.~Weinstein (1995).
\newblock Centering: A framework for modeling the local coherence of discourse.
\newblock {\em Computational Linguistics}, 21(2):203--225.

\bibitem[\protect\citeauthoryear{Hahn et~al.}{1996}]{hahn.ecai96}
Hahn,~U., K.~Markert {\&} M.~Strube (1996).
\newblock A conceptual reasoning approach to textual ellipsis.
\newblock In {\em Proc. of the 12{\/$^{th}$} European Conference on Artificial
  Intelligence (ECAI '96); Budapest, Hungary, 12--16 August 1996}, pp.
  572--576. Chichester: John Wiley.

\bibitem[\protect\citeauthoryear{Haji{\/$\check{\mbox{c}}$}ov{\'{a}}
  et~al.}{1992}]{hajicova92}
Haji{\/$\check{\mbox{c}}$}ov{\'{a}},~E., V.~Kubo{\/$\check{\mbox{n}}$} {\&}
  P.~Kubo{\/$\check{\mbox{n}}$} (1992).
\newblock Stock of shared knowledge: A tool for solving pronominal anaphora.
\newblock In {\em Proc. of the 14{\/$^{th}$} Int. Conference on Computational
  Linguistics; Nantes, France, 23-28 August 1992}, Vol.~1, pp. 127--133.

\bibitem[\protect\citeauthoryear{Kameyama}{1998}]{kameyama98}
Kameyama,~M. (1998).
\newblock Intrasentential centering: A case study.
\newblock In M.~Walker, A.~Joshi {\&} E.~Prince (Eds.), {\em Centering Theory
  in Discourse}, pp. 89--112. Oxford, U.K.: Oxford Univ. Pr.

\bibitem[\protect\citeauthoryear{Kehler}{1997}]{kehler97}
Kehler,~A. (1997).
\newblock Current theories of centering for pronoun interpretation: A critical
  evaluation.
\newblock {\em Computational Linguistics}, 23(3):467--475.

\bibitem[\protect\citeauthoryear{Lappin {\&} Leass}{1994}]{lappin94}
Lappin,~S. {\&} H.~J. Leass (1994).
\newblock An algorithm for pronominal anaphora resolution.
\newblock {\em Computational Linguistics}, 20(4):535--561.

\bibitem[\protect\citeauthoryear{Prince}{1981}]{prince81}
Prince,~E.~F. (1981).
\newblock Toward a taxonomy of given-new information.
\newblock In P.~Cole (Ed.), {\em Radical Pragmatics}, pp. 223--255. New York,
  N.Y.: Academic Press.

\bibitem[\protect\citeauthoryear{Prince}{1992}]{prince92}
Prince,~E.~F. (1992).
\newblock The {ZPG} letter: Subjects, definiteness, and information-status.
\newblock In W.~Mann {\&} S.~Thompson (Eds.), {\em Discourse Description.
  Diverse Linguistic Analyses of a Fund-Raising Text}, pp. 295--325. Amsterdam:
  John Benjamins.

\bibitem[\protect\citeauthoryear{Sidner}{1983}]{sidner83}
Sidner,~C.~L. (1983).
\newblock Focusing in the comprehension of definite anaphora.
\newblock In M.~Brady {\&} R.~Berwick (Eds.), {\em Computational Models of
  Discourse}, pp. 267--330. Cambridge, Mass.: MIT Press.

\bibitem[\protect\citeauthoryear{Strube {\&} Hahn}{1996}]{strube.acl96}
Strube,~M. {\&} U.~Hahn (1996).
\newblock Functional centering.
\newblock In {\em Proc. of the 34{\/$^{th}$} Annual Meeting of the Association
  for Computational Linguistics; Santa Cruz, Cal., 23--28 June 1996}, pp.
  270--277.

\bibitem[\protect\citeauthoryear{Suri {\&} McCoy}{1994}]{suri94}
Suri,~L.~Z. {\&} K.~F. McCoy (1994).
\newblock {RAFT/RAPR} and centering: A comparison and discussion of problems
  related to processing complex sentences.
\newblock {\em Computational Linguistics}, 20(2):301--317.

\bibitem[\protect\citeauthoryear{Walker}{1989}]{walker89}
Walker,~M.~A. (1989).
\newblock Evaluating discourse processing algorithms.
\newblock In {\em Proc. of the 27{\/$^{th}$} Annual Meeting of the Association
  for Computational Linguistics; Vancouver, B.C., Canada, 26--29 June 1989},
  pp. 251--261.

\bibitem[\protect\citeauthoryear{Walker et~al.}{1994}]{walker94}
Walker,~M.~A., M.~Iida {\&} S.~Cote (1994).
\newblock Japanese discourse and the process of centering.
\newblock {\em Computational Linguistics}, 20(2):193--233.

\end{thebibliography}

\end{document}